\documentclass[runningheads]{llncs}
\usepackage{graphicx}
\usepackage{booktabs}
\usepackage{multirow}
\usepackage{caption}
\usepackage{subcaption}
\usepackage{algorithm}
\usepackage{algorithmic}
\usepackage{amsmath,amssymb} 
\usepackage{color}
\usepackage[width=122mm,left=12mm,paperwidth=146mm,height=193mm,top=12mm,paperheight=217mm]{geometry}
\usepackage{cite}
\usepackage[colorlinks=true, linkcolor=blue, citecolor=blue, urlcolor=blue]{hyperref}
\usepackage{orcidlink}
\usepackage{fix-cm}

\usepackage{import}
\usepackage{tikz} 
\usepackage{preview} 
\PreviewEnvironment{tikzpicture} 
\usetikzlibrary{quotes,arrows.meta}
\usetikzlibrary{positioning}

\usepackage{Ball}
\usepackage{Box}
\usepackage{RightBandedBox}

\usetikzlibrary{positioning}
\usetikzlibrary{3d} 

\usepackage{xcolor}

\begin{document}
\pagestyle{headings}
\mainmatter
\def\ECCV24SubNumber{W24.11}  

\title{Optimization of Layer Skipping and Frequency Scaling for Convolutional Neural Networks under Latency Constraint} 

\titlerunning{ECCV-24 Submission ID \ECCV24SubNumber}

\authorrunning{ECCV-24 Submission ID \ECCV24SubNumber}


\makeatletter
\newcommand{\printfnsymbol}[1]{%
  \textsuperscript{\@fnsymbol{#1}}%
}
\makeatother


\author{Minh David Thao Chan\thanks{equal contribution.\\This work was supported in part by the National Natural Science Foundation of China under Grants 62341108, and in part by China Postdoctoral Science Foundation under Grant 2023M742011.}\orcidlink{0009-0000-5361-4397}
        \and Ruoyu Zhao\printfnsymbol{1}\orcidlink{0009-0008-8170-4208}
        \and Yukuan Jia\orcidlink{0000-0002-2378-9433}
        \and\\
         Ruiqing Mao\orcidlink{0000-0001-7169-3922} 
        \and Sheng Zhou\orcidlink{0000-0003-0651-0071}
}
\institute{ 
Department of Electronic Engineering, Tsinghua University, Beijing 100084, China
}

\maketitle

\begin{abstract}

The energy consumption of Convolutional Neural Networks (CNNs) is a critical factor in deploying deep learning models on resource-limited equipment such as mobile devices and autonomous vehicles. We propose an approach involving Proportional Layer Skipping (PLS) and Frequency Scaling (FS). Layer skipping reduces computational complexity by selectively bypassing network layers, whereas frequency scaling adjusts the frequency of the processor to optimize energy use under latency constraints. Experiments of  PLS and FS on ResNet-152 with the CIFAR-10 dataset demonstrated significant reductions in computational demands and energy consumption with minimal accuracy loss. This study offers practical solutions for improving real-time processing in resource-limited settings and provides insights into balancing computational efficiency and model performance.

\keywords{Layer skipping, Frequency scaling, Energy efficiency, Convolutional Neural Networks (CNNs), Autonomous vehicles}
\end{abstract}

\section{Introduction}
Convolutional Neural Networks (CNNs) have transformed fields such as computer vision and natural language processing. Deploying these computationally intensive models in resource-constrained environments such as mobile devices and autonomous vehicles presents significant challenges\cite{sudhakar2022data}. These systems require high performance, while maintaining energy efficiency and real-time constraints. For example, the perception of the environment of autonomous vehicles must satisfy strict latency requirements, reducing the battery range by up to 30\%\cite{ThaoChan2024RSU, Jia2022Online}. 
However, depending on the scenario, not all the systems are subject to the same latency constraints. Furthermore, energy-efficient techniques can drastically decrease the inference time with moderate degradation of the system performance \cite{zhang2017energy}.
Thus, understanding the trade-off between inference time, accuracy, and energy consumption is crucial for adopting efficient machine-learning applications.

This study explores a technique to address these challenges: layer skipping and frequency scaling together. The layer skipping technique selectively bypasses specific layers in sequential networks such as ResNet \cite{he2016deep}, reducing computational complexity thus shortening the inference time. However, frequency scaling reduces the computing frequency to balance performance and energy use, thereby increasing the inference time. Therefore, these two techniques should be considered together. The contributions of this study are summarized as follows:

\begin{enumerate}
    \item We propose an optimization framework involving layer skipping and frequency scaling to minimize energy consumption and computational demands in CNNs under latency constraints.
    \item We demonstrate the effectiveness of our technique on the ResNet-152 architecture with the CIFAR-10 dataset\cite{Krizhevsky2010CIFAR}, achieving significant reductions in computational demands and energy consumption while maintaining competitive accuracy both on CPU and GPU architectures.
    \item We provide insights into the trade-offs between energy efficiency and model accuracy performance, offering practical solutions for deployable models in energy-limited settings.
\end{enumerate}

\section{Related works}

The need to reduce the computation load while maintaining the performance of CNN models on mobile devices has prompted advancements in software. Model compression techniques include pruning, quantization, knowledge distillation, and low-rank factorization\cite{li2017pruningfiltersefficientconvnets,han2016deepcompressioncompressingdeep,li2017trainingquantizednetsdeeper,hinton2015distillingknowledgeneuralnetwork,ioannou2016trainingcnnslowrankfilters}. However, these methods often require customized network structures, which involve additional adjustments of the network structure for different tasks along with a longer training time, making it difficult to generalize on other CNNs and deploy on different hardware\cite{Hohman_2024}.

More general methods based on layer skipping\cite{9560049} and early exiting\cite{matsubara2022splitcomputingearlyexiting,figurnov2017spatiallyadaptivecomputationtime} have demonstrated the potential to simplify model implementation, such as BranchyNet\cite{teerapittayanon2017branchynetfastinferenceearly} and Shallow-Deep Networks\cite{kaya2019shallowdeepnetworksunderstandingmitigating} that introduce early exit branches to provide faster inference results, whereas SkipNet\cite{wang2018skipnetlearningdynamicrouting} selectively bypasses individual layers based on the output of the preceding layer, thus reducing the computation.
Other methods, such as E$^2$CM\cite{9891952}, provide an additional forward-trained network to determine the computation flow within the CNN for energy-efficient computation. However, these methods are intertwined with conditional computation where each input sample activates a different part of the model\cite{bengio2016conditionalcomputationneuralnetworks} and require joint retraining of the CNN from scratch, leading to an increase in training time and complexity.
\setlength{\emergencystretch}{3em}
From a hardware perspective, 
Dynamic Voltage and Frequency Scaling (DVFS) \cite{260129, dvfs1}
efficiently reduces the overall power consumption while maintaining the performance of the system. 
Although works investigated energy efficiency utilizing DVFS\cite{dvfs2,10.1145/3307772.3328315}, including dynamic latency constraints \cite{5090802}, very few have integrated hardware and software approaches, leaving the opportunity for further improvement in the efficiency of the system.

\section{Proposed Method }

\subsection{Preliminaries and notations}

\textbf{CNN Architecture.}
As illustrated in Fig. \ref{fig:AlexNet_Architecture}, CNNs analyze visual data through a series of layers that perform specific operations such as convolutions, activations, pooling, and fully connected transformations. Convolutional layers apply filters to extract spatial features, while activation functions like ReLU introduce non-linearity into the model. Pooling layers downsample feature maps to reduce dimensionality, enhancing computational efficiency. Fully connected layers, typically at the end of the network, produce the final classification output.Given a CNN $N$ with $m$ groups of layers, in each group $\mathcal{G}_j$ composed of residual blocs having the same spatial dimensions of feature maps and consistent filter sizes, can be defined as
\begin{equation}
    N = \bigcup_{j=1}^{m} \mathcal{G}_j,
\end{equation}
such that $\mathcal{G}_i\cap\mathcal{G}_j = \emptyset, \forall i,j \in [1..m]$, i.e., there is no overlapping between groups.    
Each group $\mathcal{G}_j$ is a subset of $N$ such that $\mathcal{G}_j = \{ \ell_{j1}, \ell_{j2}, \ldots, \ell_{jn_j} \}$ , where $\ell_{jk}$ denotes the $k$-th layer in the $j$-th group and $n_j$ is the number of layers in  the group $j$.

\begin{figure}[ht]
        \centering
        \begin{subfigure}[t]{\textwidth}
          \centering
          \includegraphics[width=0.8\linewidth]{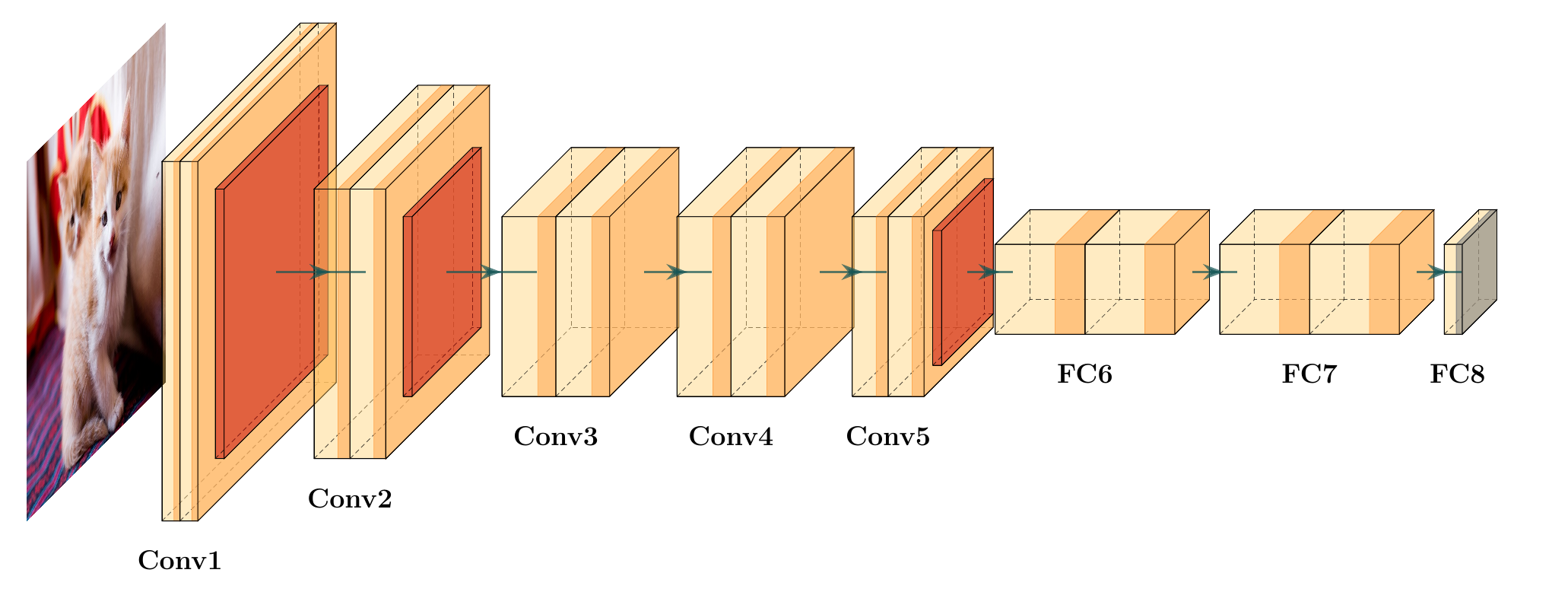}
          \setcounter{subfigure}{0}  
          \caption{AlexNet Architecture: Depicts the structure of AlexNet, showing convolutional layers (orange), pooling layers (red), and fully connected layers (grey).}
          \label{fig:AlexNet_Architecture}
        \end{subfigure}
      
        \begin{subfigure}[t]{\textwidth}
          \centering
          \includegraphics[width=0.85\linewidth]{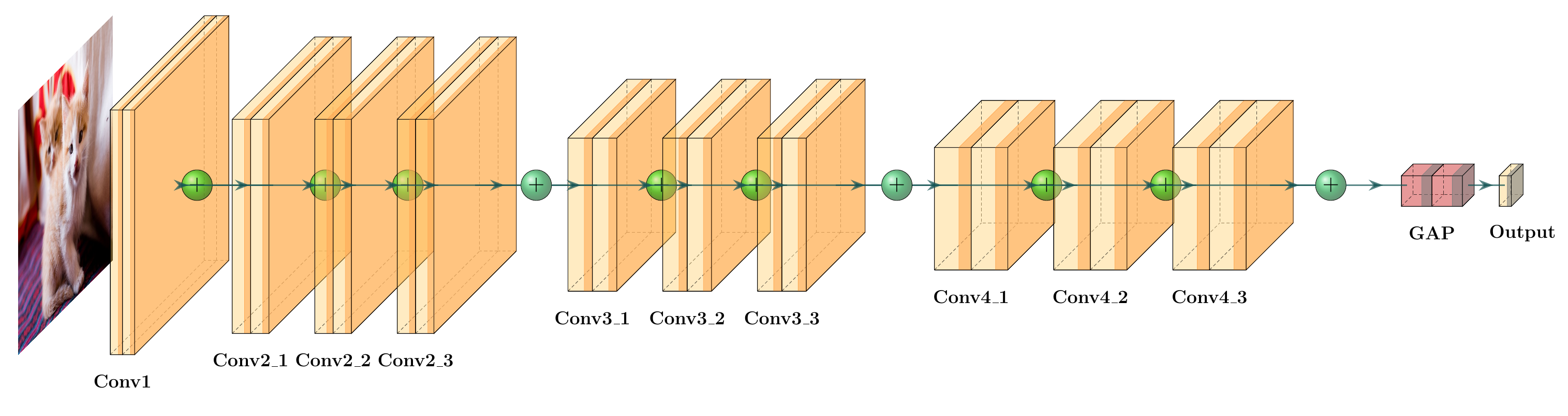}
          \setcounter{subfigure}{1}  
          \caption{ResNet Visualization with Residual Connections: Illustrates ResNet with residual connections denoted by green spheres, highlighting the skip connections across convolutional layers (orange) and the global average pooling (red) before the final output layer (grey).}
          \label{fig:ResNet_Architecture}
        \end{subfigure}

        \caption{Architectures of AlexNet (a) and ResNet (b)}
      \end{figure}

\textbf{Residual Networks}
Residual networks (ResNets) are a type of CNN architecture that leverages the skip connection mechanism, i.e. bypassing one or more layers by skipping the intermediate layers and directly adding the input of a block to its output, to improve the learning efficiency and reduce the vanishing gradient problem. 
The residual block consists of two convolutional layers, followed by a skip connection that adds the input to the output of the second convolutional layer as illustrated in Fig. \ref{fig:ResNet_Architecture}. 

Let $\mathbf{x}^i$ be the input and $F^i(\mathbf{x}^i)$ be the output of the $i^\text{th}$ layer or group of layers. Note that by construction $F^i(\mathbf{x}^i)$ and $\mathbf{x}^i$ to have the same dimensions. This requirement is satisfied by commonly used residual network architectures where
\begin{equation}
    \mathbf{x}_{\text{ResNet}}^{i+1} = F^i(\mathbf{x}_{\text{ResNet}}^i) + \mathbf{x}_{\text{ResNet}}^i,
\end{equation}

and can be addressed by pooling $\mathbf{x}^i$ to match the dimensions of $F^i(\mathbf{x}^i)$.

\textbf{Layer Skipping.}
Layer skipping allows the network to bypass some layers in the network and enter them at a later point. This technique can be used to reduce the computational complexity of the network and improve the learning efficiency. In the ResNet architecture, layer skipping is achieved by using a gating function $G^i(\mathbf{x}^i)$ that controls the flow of information between the input and the output of the $i$-th layer.
We define the output of the skipped layer (or group of layers) as
\begin{equation}
\mathbf{x}^{i+1} = G^i(\mathbf{x}^i) F^i(\mathbf{x}^i) + 
\mathbf{x}^i, 
\end{equation}
where $G^i(\mathbf{x}^i) \in \{0,1\}$ is the gating function for  the layer $i$. If $G^i(\mathbf{x}^i) = 1$, then the input is passed to the output of the $i$-th layer, and if $G^i(\mathbf{x}^i) = 0$, then the input is passed through the identity function. 

\textbf{Frequency Scaling.} 
Processors typically operate at the highest possible frequency for a given voltage. Dynamic Voltage and Frequency Scaling (DVFS) leverage the variable performance demands of computing workloads to achieve an optimal trade-off between power consumption and performance. Our focus is on frequency scaling, which dynamically adjusts the processor's operating frequency $f$ to balance energy consumption $E$ and inference time $\tau$. While increasing $f$ accelerates computations, it also raises energy usage.

The voltage-frequency (V/F) relationship defines the minimum voltage required to sustain a certain frequency. This relationship is segmented into two zones \cite{haj2018power}: a low-power zone with a constant voltage for lower frequencies, and a high-performance zone where the frequency scales approximately linearly with voltage $f \approx k \cdot V$. Consequently, in the high-performance zone, the dynamic power consumption $P_{\text{Dynamic}}$ can be approximated as:
\begin{equation}
    P_{\text{Dynamic}} = \alpha \cdot C \cdot V^2 \cdot f \approx k^2 \cdot \alpha \cdot C \cdot f^3,
\end{equation}
where $\alpha$ is a proportionality constant related to the activity factor of the processor, $C$
is the capacitance of the switched logic gates.
Therefore the total energy consumed is 
\begin{equation}
    E = \int_{0}^{\tau} (P_{\text{Dynamic}} + P_{\text{Static}}) \, dt ,  
\end{equation}

where the constant $P_{\text{{Static}}}$ is the static power consumption of the processor.
 

\textbf{Performance Metrics.}
Energy efficiency is crucial in power-constrained environments. Inference energy per frame indicates the energy used per task, essential in time-sensitive applications such as autonomous driving. Another metric is the Energy-Delay Product (EDP) which balances energy and computing latency \cite{kaxiras2008computer} given by $EDP = P \cdot t^2 = E \cdot t$, where lower values denote efficient performance with minimal energy and delay. 

We denote $a(r)$ the accuracy loss of the network $N$ when a ratio $r$ of the layers remaining. The accuracy loss is defined as the difference between the accuracy of the full network, and the accuracy of the PLS network where the network $r$ ratio of layers remain. The accuracy loss can be used as a measure performance of the network when layers are skipped. 

\textbf{Problem formulation}
The objective is to minimize the EDP  while maintaining an accuracy loss $a(r)$ over the threshold $\Gamma$ under a latency budget $d_\text{max}$. The problem $\mathcal{P}$ is formulated  as:

\begin{align}
        \mathcal{P: } & \min_{r,f} \text{ EDP} (\tau,r) \\
        &\text{{s.t. }}  \left\{
        \begin{array}{ll}
         a(r)\geq \Gamma\\
         \tau(f)\leq d_{\text{max}}\\
         f \in [f_{\text{min}}, f_{\text{max}}]\\
         r \in [r_{\text{min}},1]
        \end{array}
        \right.
\end{align}

The constraints $f \in [f_{\text{min}}, f_{\text{max}}]$ and $r \in [r_{\text{min}},1]$ define the feasible ranges for these variables while the layer skipping ratio $r$ must be within the range from $r_{\text{min}}$ to 1, where 1 indicates that no layers are skipped. These constraints ensure the optimization problem adheres to practical limits for frequency and layer skipping.

\subsection{Proportional Layer Skipping}

\begin{figure}[t]  
        \centering
        \includegraphics[width=0.9\linewidth]{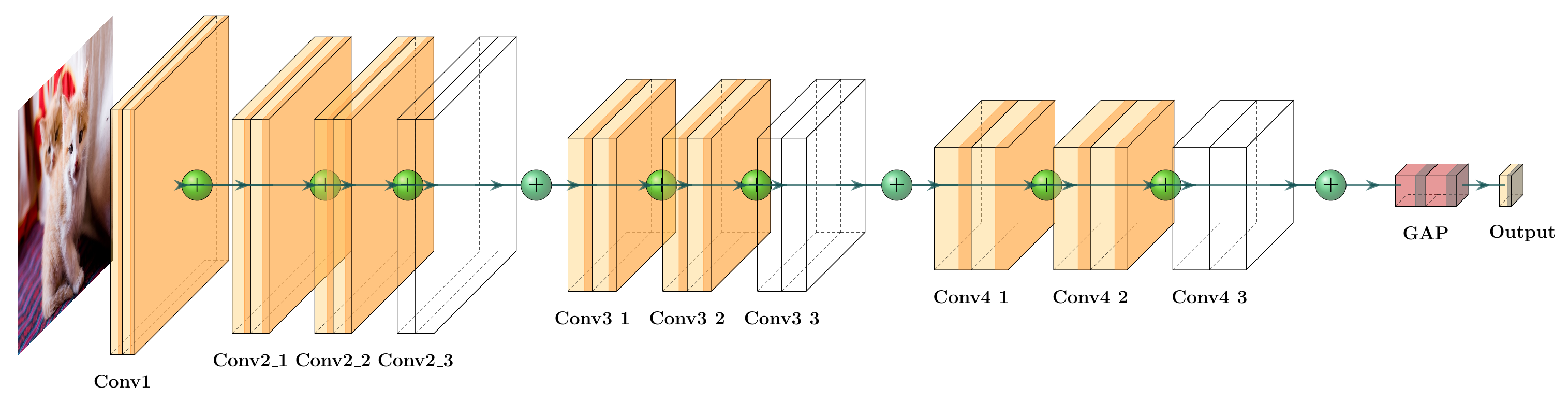}
        \setcounter{figure}{1}  
        \caption{ResNet Architecture Visualization with Residual Connections and Layer Skipping: Displays the ResNet architecture with residual connections, where green spheres represent the skip connections. Layer skipping is shown, where $r = 0.6$, indicating the proportion of layers remaining. The transition through convolutional layers (orange) to global average pooling (red) before reaching the output layer (grey) is highlighted.}
        \label{Layer_Skipping_Illustration}
\end{figure}

As shown in Fig. \ref{Layer_Skipping_Illustration} we propose a Proportional Layer Skipping (PLS) framework that enhances the computational efficiency of CNNs by using a predefined ratio $r$ to selectively skip layers. We denote PLS-$r$ the network with layer skipping ratio $r$ of remaining layers in percentage.
The Alg. \ref{PLS_Algo} involves dividing the CNN into groups of layers, $\mathcal{G}_j$, and for each group, calculating the skip start index $n_{\text{start}} = \lfloor r \cdot |\mathcal{G}_j| \rfloor$. For each layer $i$ in $\mathcal{G}_j$, the gating function $G^i(\mathbf{x}^i)$ is set to 0 if $i \geq n_{\text{start}}$ and 1 otherwise, effectively skipping the layers where $G^i(\mathbf{x}^i) = 0$.  The result is a modified network $N'$ with proportionally skipped layers, reducing computational load while maintaining accuracy. This framework is particularly suited for resource-constrained environments, such as mobile and embedded systems.

Alg. \ref{EDP_Algo} optimizes energy efficiency by selecting the largest layer skipping ratio \(r\) for a desired accuracy \(\Gamma\), then choosing a processor frequency \(f\) that meets the latency constraint \(d_{\text{max}}\) while minimizing the Energy-Delay Product (EDP).


\begin{figure}[t]
        \centering
        \begin{minipage}[t]{0.48\textwidth}
            \centering
            \begin{algorithm}[H]
            \caption{Proportional \\ Layer Skipping}
            \begin{algorithmic}[1]
            \REQUIRE CNN $N$, ratio $r$
            \ENSURE Modified CNN $N'$ with skipped layers
            \FOR{each group of layers $\mathcal{G}_j$ in $N$}
            \STATE $n_{\text{start}} = \lfloor r \cdot |\mathcal{G}_j| \rfloor$
            \FOR{each layer $i$ in $\mathcal{G}_j$}
                    \STATE $G^i(\mathbf{x}^i) = \begin{cases}
                    0 & \text{if } i \geq n_{\text{start}} \\
                    1 & \text{otherwise}
                    \end{cases}$
            \ENDFOR
            \ENDFOR
            \RETURN  $N'$
            \end{algorithmic}
                    \label{PLS_Algo}
            \end{algorithm}
        \end{minipage}%
        \hfill
        \begin{minipage}[t]{0.48\textwidth}
            \centering
            \begin{algorithm}[H]
                    \caption{Frequency and Layer Skipping Ratio Selection}
                    \begin{algorithmic}[1]
                    \REQUIRE $d_{\text{max}}$, $\Gamma$
                    \ENSURE Optimal $f$, $r$
                    \STATE Choose the largest $r$ such that $a_r \geq \Gamma$
                    \STATE Select $f$ such that $\tau(f) \leq d_{\text{max}}$ and minimizes EDP
                    \RETURN $f$, $r$
                    \end{algorithmic}
                    \label{EDP_Algo}
            \end{algorithm}
        \end{minipage}
    \end{figure}

\section{Experiments}

We deployed the PLS algorithm on Nvidia Jetson Orin equipped with 6 CPUs and 1 GPU, to selectively skip layers in ResNet-152 and evaluated the performance of PLS-$r$ on the CIFAR-10 dataset\cite{Krizhevsky2010CIFAR}. The DVFS is tested for designated values of frequencies. The network number of parameters and accuracy before and PLS are shown in Table \ref{Table1}.

\begin{table}[ht]
        \centering
        \caption{Comparison of ResNet-152 Before and After Layer Skipping}
        \label{Table1}
        \begin{tabular}{lccc}
        \toprule
        \textbf{Network} & \textbf{Accuracy} & \textbf{\# Parameters} & \textbf{Compression Rate} \\
        \midrule
        ResNet-152 Reference & 87.64\% & 58.16M & - \\
        ResNet-152 PLS-90 & 86.16\% & 54.81M & 5.76\% \\
        ResNet-152 PLS-80 & 83.96\% & 50.06M & 13.93\% \\
        ResNet-152 PLS-70 & 81.18\% & 46.42M & 20.19\% \\
        ResNet-152 PLS-50 & 77.87\% & 29.04M & 50.07\% \\
        ResNet-152 PLS-30 & 69.05\% & 17.25M & 70.34\% \\
        ResNet-152 PLS-10 & 55.65\% & 9.15M & 84.27\% \\
        \bottomrule
        \end{tabular}
        \end{table}
        This table highlights the trade-off in ResNet-152 between layer skipping and performance. As layers are skipped, the model compresses, with accuracy gradually decreasing. The remaining 50\% of layers cuts parameters by half, but accuracy drops to 77.87\%. The remaining 10\% results in significant compression (84.27\%) but a steep accuracy decline to 55.65\%. This demonstrates the balance between model efficiency and accuracy.

\subsection{Ablation Study on layers in groups}
In this ablation study, we investigate the impact of layer skipping and varying dataset sizes on the performance of the ResNet-152 model. The primary objective is to understand how these factors influence accuracy, model complexity, and inference time, providing insights for optimizing computational efficiency while maintaining acceptable performance levels. In particular, we aim to determine how layer skipping and dataset size affect the accuracy and the trade-offs between computational efficiency and accuracy, especially in resource-constrained environments.
\begin{figure}[ht]
        \centering
        \begin{subfigure}{.32\textwidth}
          \centering
          \includegraphics[width=\linewidth]{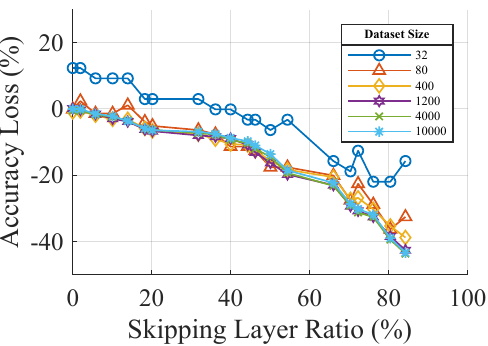}
          \caption{}
          \label{3_Abla_Accuracy_Testsize}
        \end{subfigure}%
        \hfill
        \begin{subfigure}{.32\textwidth}
          \centering
          \includegraphics[width=\linewidth]{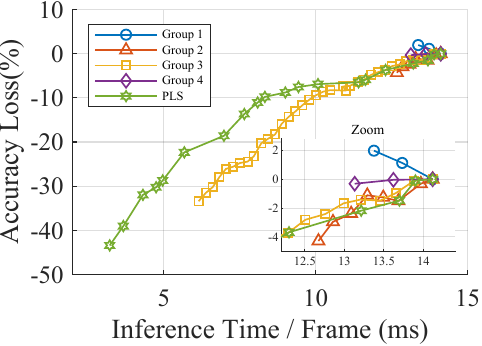}
          \caption{}
          \label{2_Abla_Accuracy_Latency}
        \end{subfigure}%
        \hfill
        \begin{subfigure}{.32\textwidth}
          \centering
          \includegraphics[width=\linewidth]{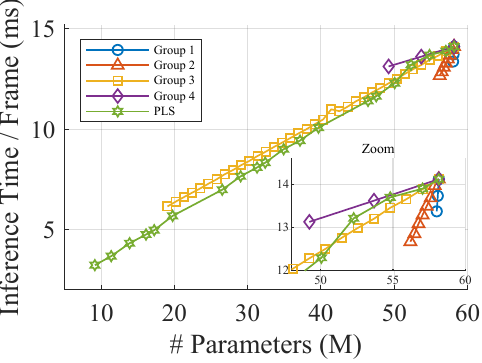}
          \caption{}
          \label{0_Abla_Latency_Params}
        \end{subfigure}
        \caption{Impact of Layer Skipping and Dataset Size on ResNet-152:
        (a) Accuracy vs. Parameters: Shows accuracy loss with reduced parameters due to layer skipping.
        (b) Accuracy vs. Inference Time: Illustrates trade-off between faster inference and accuracy.
        (c) Inference Time vs. Complexity: Linear relationship between reduced complexity and faster inference.}
        \label{fig:Impact of Layer Skipping and Dataset Size on ResNet-152}
      \end{figure}

      \textbf{Methodology.}
      We conducted experiments on the CIFAR-10 dataset\cite{Krizhevsky2010CIFAR} using ResNet-152, varying the dataset sizes from 32 to 10,000 to measure accuracy. The network layers were grouped, and layers within each group were progressively skipped according to the proposed PLS framework. We analyzed the impact on three key metrics: accuracy loss, number of parameters, and inference time per frame.  Layers were grouped based on their relative depth within the network, with early layers grouped separately from deeper layers to examine their distinct impact.
      
      \textbf{Results analysis and discussion.}
      As shown in Fig. \ref{fig:Impact of Layer Skipping and Dataset Size on ResNet-152}, several trends emerge.
      Accuracy decreases as the proportion of skipped layers increases, for a dataset sufficiently large as shown in Fig. \ref{3_Abla_Accuracy_Testsize}. 
      In Fig. \ref{2_Abla_Accuracy_Latency}, as layers are skipped, inference time decreases, but this comes at the cost of accuracy. The PLS framework, highlighted by the green line shows a relatively balanced trade-off. The larger the group is (i.e. the more points), the more removing a layer decreases accuracy. The accuracy of group 3 composed of 32 layers is below that of group 2 composed of 3 layers.
      In Fig. \ref{0_Abla_Latency_Params}, the number of parameters decreases linearly with the inference time. The later the group, the more removing one layer from that group reduces the number of parameters. For example, removing a layer from group 4 reduces by 10 M the number of parameters whereas removing a layer from group 2 reduces it by 1 M. Consequently, the later the group
     the more removing its layer reduces the inference time.

\subsection{GPU hardware architecture in linear region}

\textbf{Methodology.}  We conducted experiments on an Nvidia Jetson Orin GPU to analyze the effects of frequency scaling and layer skipping. The PLS algorithm was tested at typical frequencies ranging from 306 MHz to 624.75 MHz. The setup involved isolating GPU metrics by assigning non-inference tasks to the CPU, ensuring that the impact on GPU performance was accurately measured.
\begin{figure}[ht]
        \centering
        \begin{subfigure}[t]{.24\textwidth}
            \centering
            \includegraphics[width=\linewidth]{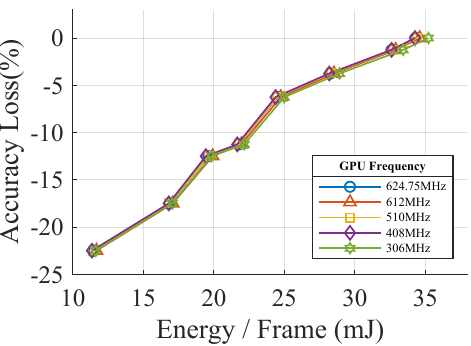}
            \caption{}
            \label{4_GPU_Accuracy_Energy}
        \end{subfigure}%
        \hfill
        \begin{subfigure}[t]{.24\textwidth}
            \centering
            \includegraphics[width=\linewidth]{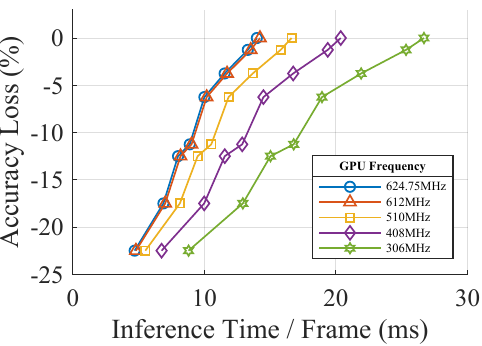}
            \caption{}
            \label{5_GPU_Accuracy_Latency}
        \end{subfigure}%
        \hfill
        \begin{subfigure}[t]{.24\textwidth}
            \centering
            \includegraphics[width=\linewidth]{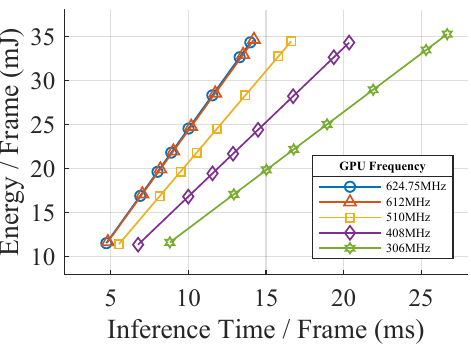}
            \caption{}
            \label{6_GPU_Energy_Latency_freq}
        \end{subfigure}%
        \hfill
        \begin{subfigure}[t]{.24\textwidth}
            \centering
            \includegraphics[width=\linewidth]{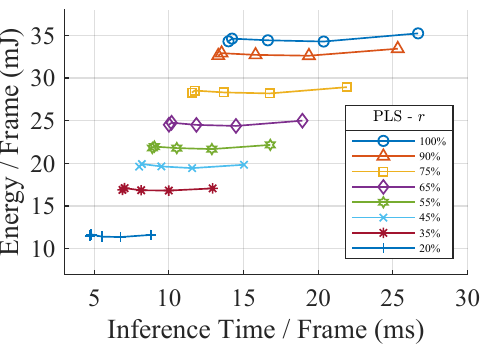}
            \caption{}
            \label{7_GPU_Energy_Latency_layer}
        \end{subfigure}
        \caption{Impact of Frequency Scaling and Layer Skipping on GPU Performance. (a) Trade-off between Accuracy Loss and Energy per Frame. (b) Trade-off between Accuracy Loss and Inference Time per Frame. (c) Relationship between Energy per Frame and Inference Time per Frame under Frequency Scaling. (d) Energy-Delay Product Analysis across Different Layer Skipping Ratios.
        }
        \label{fig:Impact of Frequency Scaling and Layer Skipping on GPU Performance}
    \end{figure}

\textbf{Results Analysis and Discussion.} As shown in Fig. \ref{fig:Impact of Frequency Scaling and Layer Skipping on GPU Performance}, energy consumption decreased significantly with PLS. With a 23\% loss in accuracy compared to the full ResNet-152, the energy consumption decreased from 35 mJ to 11 mJ, representing a 69\% reduction in energy per frame, highlighting the potential for significant energy savings with minimal compromise in accuracy. Furthermore, the accuracy and energy are only dependent on the ratio $r$ chosen of PLS, i.e., frequency scaling does not have an impact on the inference energy/frame as shown in Fig. \ref{4_GPU_Accuracy_Energy}.

However, frequency scaling directly influences inference time. Reducing the frequency from 612 MHz to 306 MHz doubled the inference time for both full ResNet-152 and PL20\% configurations. Specifically, the average inference time per frame increases from 13 ms to 26.5 ms and from 4.5 ms to 9 ms, respectively as shown in Fig. \ref{5_GPU_Accuracy_Latency}. This demonstrates an inversely proportional relationship between the frequency and inference time. 

In Fig. \ref{6_GPU_Energy_Latency_freq}, for the same frequency, the energy consumption exhibited a linear relationship with inference time, indicating constant power consumption. Higher frequencies resulted in steeper tangents, reflecting increased power usage. This linearity confirms that power is constant for specific frequencies. A direct proportional relationship between frequency scaling and inference time is evident. Reducing the frequency by half from 612 MHz to 306 MHz doubles the average inference time per frame from 13 ms to 26.5 ms and from 4.5 ms to 9 ms for the full ResNet-152 and PL20\%, respectively. 

Iso-accuracy curves in Fig. \ref{7_GPU_Energy_Latency_layer} are parallel and exhibit constant energy consumption, indicating that energy use is solely determined by the layer skipping ratio $r$ of the PLS for a given accuracy level. This consistency suggests that within this configuration, the energy consumption remains fixed for a given model, regardless of the frequency scaling.

\subsection{CPU Hardware Architecture in cubic region}
\textbf{Methodology.}In the second part of our study, we focused on CPU performance for inference, testing our PLS algorithm at frequencies ranging from 268.8 MHz to 1.51 GHz on 2 CPUs.

\begin{figure}[ht]
        \centering
        \begin{subfigure}[t]{.24\textwidth}
            \centering
            \includegraphics[width=\linewidth]{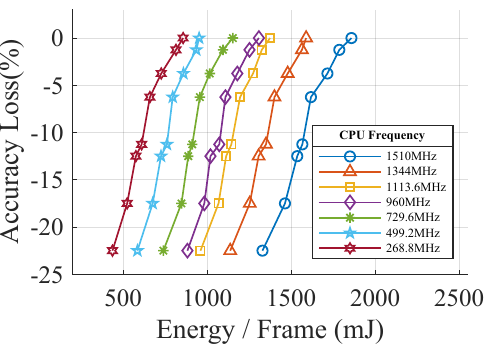}
            \caption{}
            \label{9_CPU_Accuracy_Energy}
        \end{subfigure}%
        \hfill
        \begin{subfigure}[t]{.24\textwidth}
            \centering
            \includegraphics[width=\linewidth]{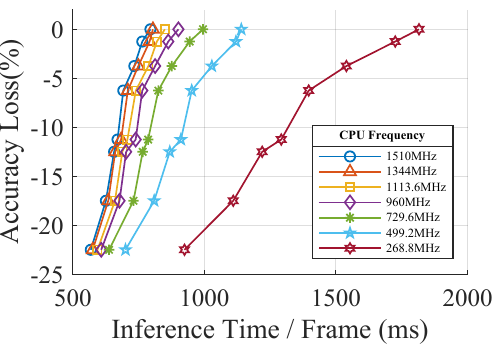}
            \caption{}
            \label{10_CPU_Accuracy_Latency}
        \end{subfigure}%
        \hfill
        \begin{subfigure}[t]{.24\textwidth}
            \centering
            \includegraphics[width=\linewidth]{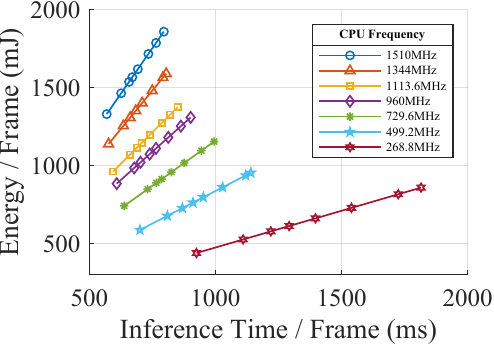}
            \caption{}
            \label{11_CPU_Energy_Latency_freq}
        \end{subfigure}%
        \hfill
        \begin{subfigure}[t]{.24\textwidth}
            \centering
            \includegraphics[width=\linewidth]{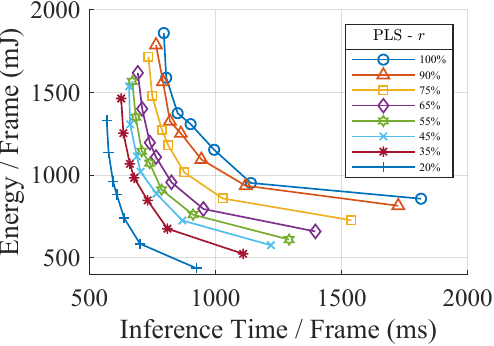}
            \caption{}
            \label{12_CPU_Energy_Latency_layer}
        \end{subfigure}
        \caption{CPU Analysis of frequency scaling and layer skipping effects. (a) Accuracy Loss vs. Energy per Frame. (b) Accuracy Loss vs. Inference Time per Frame. (c) Energy per Frame vs. Inference Time per Frame across CPU frequencies. (d) Energy per Frame vs. Inference Time per Frame for different layer skipping ratios.}
\end{figure}

\textbf{Results Analysis and Discussion.} As illustrated in Fig. \ref{9_CPU_Accuracy_Energy}, reducing the CPU frequency leads to a significant decrease in energy consumption for maintaining the same accuracy level. Specifically, lowering the frequency from 1.51 GHz to 729 MHz for the full ResNet-152 model results in a 36\% reduction in average inference energy consumption. However, this energy efficiency gain comes at the cost of an inversely proportional increase in inference time, as demonstrated in Fig. \ref{10_CPU_Accuracy_Latency}. This trade-off highlights the need for balancing performance with energy efficiency, especially in scenarios where computational resources are limited.

In Fig. \ref{11_CPU_Energy_Latency_freq}, similarly to the GPU, a linear relationship between energy consumption and inference time at a constant frequency is observed, indicating consistent power usage across different operational conditions. Higher frequencies exhibit steeper slopes, reflecting increased power consumption. 

Furthermore, Fig. \ref{12_CPU_Energy_Latency_layer} provides additional insight into the interplay between accuracy and energy efficiency. For the same accuracy level, longer inference times correspond to lower energy usage, particularly when the frequency is reduced from 1.51 GHz to 499.2 MHz. This observation suggests that for tasks where inference time is less critical, reducing frequency can yield substantial energy savings. Nonetheless, achieving higher accuracy consistently demands more energy, regardless of the inference frequency.

\subsection{Energy and EDP Analysis of the PLS Algorithm}

\begin{figure}[ht]
        \centering
        \begin{subfigure}[t]{.24\textwidth}
            \centering
            \includegraphics[width=\linewidth]{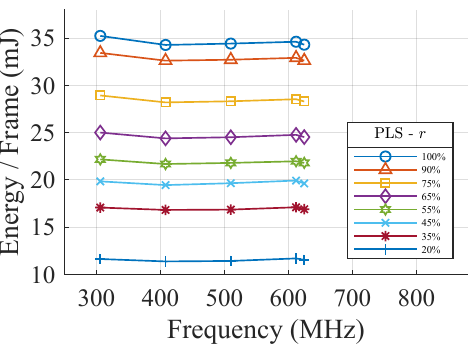}
            \caption{}
            \label{8_GPU_Energy_Freq}
        \end{subfigure}%
        \hfill
        \begin{subfigure}[t]{.24\textwidth}
            \centering
            \includegraphics[width=\linewidth]{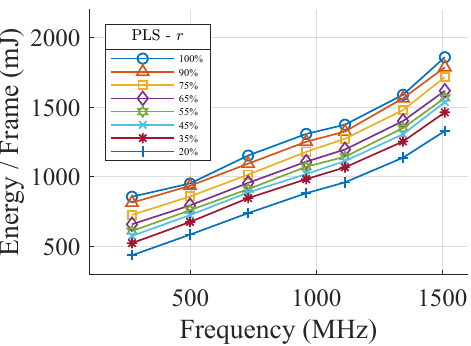}
            \caption{}
            \label{14_CPU_Energy_Freq}
        \end{subfigure}%
        \hfill
        \begin{subfigure}[t]{.24\textwidth}
            \centering
            \includegraphics[width=\linewidth]{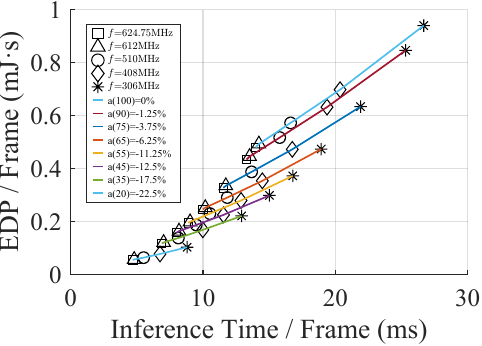}
            \caption{}
            \label{16_GPU_EDP_Latency_freq}
        \end{subfigure}%
        \hfill
        \begin{subfigure}[t]{.24\textwidth}
            \centering
            \includegraphics[width=\linewidth]{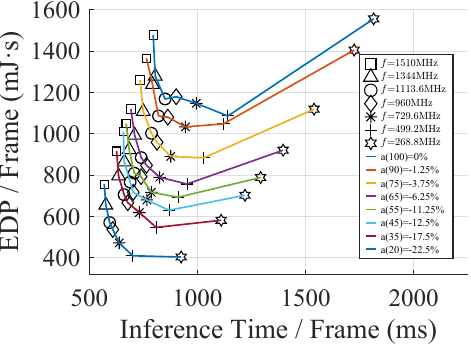}
            \caption{}
            \label{18_CPU_EDP_Latency_freq}
        \end{subfigure}
        \caption{Impact of Frequency Scaling and Layer Skipping on Energy Efficiency and EDP. (a) GPU energy consumption across different frequencies, showing relative insensitivity to scaling. (b) CPU energy consumption increases linearly below 1 GHz, with a steeper rise at higher frequencies. (c) EDP vs. inference time on GPU, demonstrating the trade-off between energy efficiency and latency. (d) EDP vs. inference time on CPU, highlighting optimal frequency combinations for minimizing EDP across different layer skipping ratios and frequencies.}
    \end{figure}

        
        
        

We evaluated inference energy efficiency using two key metrics: average energy per frame and EDP per frame. As depicted in Fig. \ref{8_GPU_Energy_Freq}, GPU energy consumption remains relatively constant across different frequencies, indicating limited sensitivity to scaling. Conversely, Fig. \ref{14_CPU_Energy_Freq} reveals that CPU energy consumption increases linearly at frequencies below 1 GHz, followed by a steeper rise at higher frequencies. 
This pattern indicates that the CPU’s energy usage is more sensitive to frequency scaling therefore giving more opportunity for energy saving.

When considering EDP, which balances energy consumption and inference time, we identified an optimal frequency that minimizes EDP while tolerating a slight accuracy loss. For GPUs, EDP is minimized at the shortest inference time corresponding to a given accuracy level as shown in Fig. \ref{16_GPU_EDP_Latency_freq}. In contrast, for CPUs, an optimal frequency that balances these factors is demonstrated in Fig. \ref{18_CPU_EDP_Latency_freq}  with the optimal EDP for the original ResNet-152 achieved at 499 MHz.

\section{Conclusions}

We presented a study investigating the trade-off between energy efficiency and model accuracy by skipping layers and frequency scaling. Our study was motivated by the limited computing and energy resources on mobile systems under time and accuracy constraints, such as autonomous vehicles, by skipping layers and adjusting the accuracy of the system without the need for retraining. We highlight our experiments on CIFAR-10 for CPU and GPU architectures, which show that ResNet-152 can be proportionally skipped while striking a balance between energy efficiency and accuracy.
This work leads to mobile computing devices with latency constraints and energy requirements for real-time processing, making deployment on all CPU, GPU, or mixed architectures in the real world easier.

\bibliographystyle{splncs}
\end{document}